\pdfoutput=1
\documentclass[11pt]{article}

\usepackage{acl}

\usepackage{times}
\usepackage{latexsym}

\usepackage[T1]{fontenc}
\usepackage[utf8]{inputenc}

\usepackage{microtype}

\usepackage{inconsolata}

\usepackage{amsmath}
\usepackage{amssymb}
\usepackage{graphicx}
\usepackage{tabularx}
\usepackage{multirow}
\usepackage{multicol}
\usepackage[justification=justified,singlelinecheck=false]{caption}
\usepackage{ragged2e}
\usepackage[ruled,vlined,linesnumbered]{algorithm2e}
\usepackage{verbatim}
\usepackage{booktabs}
\usepackage{fontawesome5}
\usepackage{subcaption}
\usepackage[draft]{minted}
\usepackage{hyperref}
\usepackage[noabbrev,capitalize,nameinlink]{cleveref}
\usepackage{tikz}
\usepackage{tikz-dependency}
\usepackage{adjustbox}
\usepackage{todonotes}
\usepackage[multiple]{footmisc}
\usepackage{pgfplots}
\usepackage{rotating}
\usetikzlibrary{positioning, fadings}
\usepackage{transparent}
\usepackage{cclicenses}
\usepackage[
ragged
]{sidecap}
\usepackage{bbm}
\usepackage{colortbl}
\usepackage{adjustbox}

\newcommand{\knn}{$k$NN}
\newcommand{\nnose}{NNOSE}
\newcommand*\circled[1]{\tikz[baseline=(char.base)]{
            \node[shape=circle,draw,inner sep=.6pt] (char) {#1};}}

\newcommand{\vanilla}{\multirow{3}{*}{\rotatebox[origin=c]{90}{Vanilla}}}
\newcommand{\withknn}{\multirow{3}{*}{\rotatebox[origin=c]{90}{+\knn{}}}}
\newcommand{\dd}{\texttt{\{D\}}}
\newcommand{\dwt}{\texttt{{\{D\}}+WT}}
\newcommand{\ad}{$\forall$\texttt{D}}
\newcommand{\adwt}{$\forall$\texttt{D+WT}}

\newcommand{\bertb}{BERT\textsubscript{base}}
\newcommand{\roberta}{RoBERTa\textsubscript{base}}
\newcommand{\up}[1]{{\color{blue} $\uparrow$\textsubscript{#1}}}
\newcommand{\down}[1]{{\color{red} $\downarrow$\textsubscript{#1}}}
\newcommand{\none}[1]{{--\textsubscript{#1}}}
\usepackage{todonotes}
\definecolor{kmy-color}{rgb}{0.858, 0.188, 0.478}

\title{NNOSE: {N}earest {N}eighbor {O}ccupational {S}kill {E}xtraction}

\author{Mike Zhang\textsuperscript{\faCompass}\textsuperscript{\faRobot} \hspace{1em}
Rob van der Goot\textsuperscript{\faCompass}\textsuperscript{\faRobot} \hspace{1em}
Min-Yen Kan\textsuperscript{\faGlobe} \hspace{1em}
Barbara Plank\textsuperscript{\faCompass}\textsuperscript{\faMountain}\textsuperscript{\faHiking}\\
\textsuperscript{\faCompass}Department of Computer Science, IT University of Copenhagen, Denmark\\
\textsuperscript{\faRobot}Pioneer Centre for Artificial Intelligence, Copenhagen, Denmark \\
\textsuperscript{\faGlobe}School of Computing, National University of Singapore, Singapore \\
\textsuperscript{\faMountain}MaiNLP, Center for Information and Language Processing, LMU Munich, Germany \\
\textsuperscript{\faHiking}Munich Center for Machine Learning (MCML), Munich, Germany \\
{\tt mikejj.zhang@gmail.com }}

\begin{document}
\maketitle
\begin{abstract}
The labor market is changing rapidly, prompting increased interest in the automatic extraction of occupational skills from text. 
With the advent of English benchmark job description datasets, there is a need for systems that handle their diversity well. 
We tackle the complexity in occupational skill datasets tasks---combining and leveraging multiple datasets for skill extraction, to identify rarely observed skills within a dataset, and overcoming the scarcity of skills across datasets.
In particular, we investigate the retrieval-augmentation of language models, employing an external datastore for retrieving similar skills in a dataset-unifying manner. 
Our proposed method, \textbf{N}earest \textbf{N}eighbor \textbf{O}ccupational \textbf{S}kill \textbf{E}xtraction (NNOSE) effectively leverages multiple datasets by retrieving neighboring skills from other datasets in the datastore.  
This improves skill extraction \emph{without} additional fine-tuning. 
Crucially, we observe a performance gain in predicting infrequent patterns, with substantial gains of up to 30\% span-F1 in cross-dataset settings. 
\end{abstract}

\section{Introduction}
Labor market dynamics, influenced by technological changes, migration, and digitization, have led to the availability of job descriptions (JD) on platforms to attract qualified candidates~\cite{brynjolfsson2011race,brynjolfsson2014second,balog2012expertise}. 
JDs consist of a collection of skills that exhibit a characteristic \emph{long-tail pattern}, where popular skills are more common while niche expertise appears less frequently across industries~\cite{autor2003skill,autor2013growth}, such as ``teamwork'' vs.\ ``system design''.\footnote{Examples are from the \href{https://rb.gy/3zgld}{\texttt{CEDEFOP Skill Platform}}.} 
This pattern poses challenges for skill extraction (SE) and analysis, as certain skills may be underrepresented, overlooked, or emerging in JDs.
This complexity makes the extraction and analysis of skills more difficult, resulting in a \emph{sparsity of skills} in SE datasets. We tackle this by combining three different skill datasets.

To address the challenges in SE, we explore the potential of Nearest Neighbors Language Models (NNLMs;~\citealp{Khandelwal2020Generalization}). NNLMs calculate the probability of the next token by combining a parametric language model (LM) with a distribution derived from the k-nearest context--token pairs in the datastore. This enables the storage of large amounts of training instances without the need to retrain the LM weights, improving language modeling. However, the extent to which NNLMs enhance application-specific end-task performance beyond language modeling remains relatively unexplored. Notably, NNLMs offer several advantages, as highlighted by~\citet{Khandelwal2020Generalization}: 
First, explicit memorization of the training data aids generalization.
Second, a single LM can adapt to multiple domains without domain-specific training, by incorporating domain-specific data into the datastore (e.g., multiple datasets).
Third, the NNLM architecture excels at predicting rare patterns, particularly the long-tail.

\begin{SCfigure*}
    \centering
    \includegraphics[width=10cm]{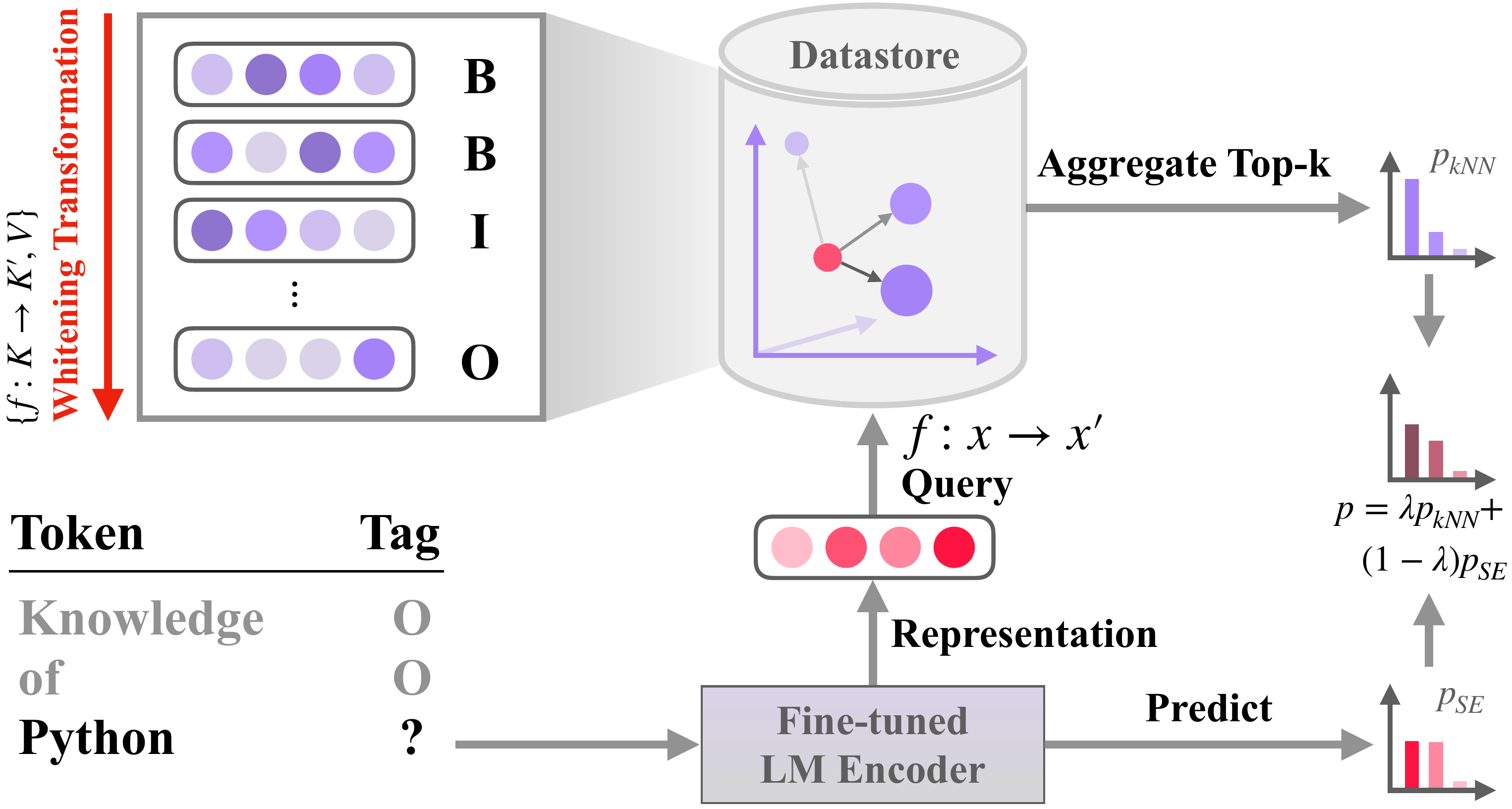}
    \caption{
    \justifying\textbf{Setup of \nnose{}.} The datastore consists of paired contextual token representations obtained from a fine-tuned encoder and the corresponding \texttt{BIO} tag. 
    We use a whitening transformation to enhance the isotropy of token representations. 
    During inference, i.e., retrieving tokens, we use the same whitening transformation on the test token’s representation to retrieve the $k$-nearest neighbors from the datastore. 
    We interpolate the encoder and \knn{} distributions with a hyperparameter $\lambda$ as the final distribution.
    }
    \label{fig:fig1}
\end{SCfigure*}

Therefore, we seek to answer the question:\ \emph{\textbf{How effective are nearest neighbors retrieval methods for occupational skill extraction?}} 
Our contributions are as follows:
\begin{itemize}
\itemsep0em
    \item To the best of our knowledge, we are the first to investigate encoder-based \knn{} retrieval by leveraging \emph{multiple} datasets.
    \item Furthermore, we present a novel domain-specific \roberta{}-based language model, JobBERTa, tailored to the job market domain.
    \item We conduct an extensive analysis to show the advantages of \knn{} retrieval, in contrast to prior work that primarily focuses on hyperparameter-specific analysis.\footnote{Code and data: \url{https://github.com/mainlp/nnose}.}
\end{itemize}

\section{Nearest Neighbor Skill Extraction}
\paragraph{Skill Extraction.} The task of SE is formulated as a sequence labeling problem. 
We define a set of job description sentences $\mathcal{X}$, where each $d \in \mathcal{X}$ represents a set of sequences with the $j^{\text{th}}$ input sequence $\mathcal{X}^j_d= \{x_1, x_2, ..., x_i\}$, with a corresponding target sequence of \texttt{BIO}-labels $\mathcal{Y}^j_d = \{y_1, y_2, ..., y_i\}$. The labels include ``\texttt{B}'' (beginning of a skill token), ``\texttt{I}'' (inside skill token), and ``\texttt{O}'' (any outside token). 
The objective is to use $\mathcal{D}$ in training a labeling algorithm that accurately predicts entity spans by assigning an output label $y_i$ to each token $x_i$.

\subsection{\nnose{}}\label{subsec:knnse}
\looseness=-1
The core idea of \nnose{} is that we augment the extraction of skills during inference with a \knn{} retrieval component and a datastore consisting of context--token pairs. 
\cref{fig:fig1} outlines our two-step approach. 
First, we extract skills by getting token representation $\boldsymbol{h}_i$ from $x_i$ and assign a probability distribution $p_{\mathrm{SE}}$ for each $\boldsymbol{h}_i$ in the input sentence. 
Second, we use each $\boldsymbol{h}_i$ to find the most similar token representations in the datastore and get the probability distribution $p_{\mathrm{kNN}}$, aggregated from the $k$-nearest context--token pairs. 
Last, we obtain the final probability distribution $p$ by interpolating between the two distributions. 
In addition to formalizing \nnose{}, we apply the Whitening Transformation (\cref{subsec:zca}) to the embeddings, an important process for \knn{} approaches as used in previous work~\cite{su2021whitening, yin2022efficient}.

\paragraph{Datastore.} 
The datastore $\mathcal{D}$ comprises key--value pairs $(\boldsymbol{h}_i, y_i)$, where each $\boldsymbol{h}_i$ represents the contextualized token embedding computed by a \emph{fine-tuned} SE encoder, and $y_i \in \{\text{\tt{B}, \tt{I}, \tt{O}}\}$ denotes the corresponding gold label. 
Typically, the datastore consists of all tokens from the training set. 
In contrast to the approach employed by~\citet{wang2022k} for \knn{}--NER, where they only store \texttt{B} and \texttt{I} tags in the datastore (only named entities), we also include the \texttt{O}-tag in the datastore. This allows us to retrieve non-named entities, which is more intuitive than assigning non-entity probability mass to the \texttt{B} and \texttt{I} tokens.

\paragraph{Inference.} 
During inference, the \nnose{} model aims to predict $y_i$ based on the contextual representation of $x_i$ (i.e., $\boldsymbol{h}_i$). 
This representation is used to query the datastore for \knn{} using an $L^2$ distance measure (following~\citealp{Khandelwal2020Generalization}), denoted as $d(\cdot, \cdot)$. 
Once the neighbors are retrieved, the model computes a distribution over the neighbors by applying a softmax function with a temperature parameter $T$ to their negative distances (i.e., similarities). 
This aggregation of probability mass for each label (\text{\tt{B}, \tt{I}, \tt{O}}) across all occurrences in the retrieved targets is represented as:

\begin{equation}
\small
p_{\mathrm{kNN}}(y_i \mid x_i) \propto \sum_{(k_i, v_i) \in \mathcal{D}} \mathbbm{1}_{y=v_i} \exp\left(\frac{-d(\boldsymbol{h}_i, \boldsymbol{k})}{T}\right).
\end{equation}

Items that do not appear in the retrieved targets have zero probability. 
Finally, we interpolate the nearest neighbors distribution $p_{\mathrm{kNN}}$ with the fine-tuned model distribution $p_{\mathrm{SE}}$ using a tuned parameter $\lambda$ to produce the final \nnose{} distribution $p$:

\begin{equation}
\begin{aligned}
p(y_i \mid x_i) = \hspace{1mm} & \lambda \times p_{\mathrm{kNN}}\left(y_i \mid x_i\right) + \\ & (1-\lambda) \times p_{\mathrm{SE}}\left(y_i \mid x_i\right).
\end{aligned}
\end{equation}

\begin{table*}[t]
\centering
\resizebox{.8\linewidth}{!}{
\begin{tabular}[t]{lllrrrr}
\toprule
\textbf{Dataset}       &    \textbf{Location}    & \textbf{License}    &   \textbf{Train}   & \textbf{Dev.}  &   \textbf{Test}  & $\mathcal{D}$ (\textbf{Tokens}) \\\midrule
\textsc{SkillSpan}     &    *        & CC-BY-4.0  &   5,866   & 3,992 &   4,680 & 86.5K  \\
\textsc{Sayfullina}    &    UK       & Unknown    &   3,706   & 1,854 &   1,853 & 53.1K   \\
\textsc{Green}         &    UK       & CC-BY-4.0  &   8,670   & 963   &   336   & 209.5K  \\\midrule
\textsc{Total}         & \multicolumn{5}{c}{}                                                          & 349.2K \\
\bottomrule
\end{tabular}}
\caption{\textbf{Dataset Statistics.} We provide statistics for all three datasets, including the location and license. Input granularity is at the token level, with performance measured in span-F1. The size of the datastore $\mathcal{D}$ is in tokens and determined by embedding tokens and their context from the training sets, resulting in approximately 350K keys. See~\cref{data:examples} for examples.
}
\label{tab:num_post}
\end{table*}

\subsection{Whitening Transformation}\label{subsec:zca}
Several works (\citealp{li-etal-2020-sentence, su2021whitening, huang-etal-2021-whiteningbert-easy}) note that if a set of vectors are isotropic, we can assume it is derived from the Standard Orthogonal Basis, which also indicates that we can properly calculate the similarity between embeddings. 
Otherwise, if it is anisotropic, we need to transform the original sentence embedding to enforce isotrophorism, and then measure similarity. \citet{su2021whitening, huang-etal-2021-whiteningbert-easy} applies the vector whitening approach~\cite{koivunen1999feasibility} on BERT~\cite{devlin2019bert}. 
The Whitening Transformation (\texttt{WT}), initially employed in data preprocessing, aims to eliminate correlations among the input data features for a model.
In turn, this can improve the performance of certain models that rely on uncorrelated features.
Other works (\citealp{gao2018representation, ethayarajh2019contextual, li2020sentence, yan2021consert, jiang2022promptbert}, among others) found that (frequency) biased \emph{token} embeddings hurt final sentence representations. 
These works often link token embedding bias to the token embedding anisotropy and argue it is the main reason for the bias. 
We apply \texttt{WT} to the token embeddings like previous work for nearest neighbor retrieval~\cite{yin2022efficient}. 
In short, \texttt{WT} transforms the mean value of the embeddings into 0 and the covariance matrix into the identity matrix, and these transformations are then applied to the original embeddings. 
We apply \texttt{WT} to the embeddings before putting them in the datastore and before querying the datastore.
The workflow of \texttt{WT} is detailed in \cref{app:wt}.

\section{Experimental Setup}\label{sec:exp}

\subsection{Data}

All datasets are in English and have different label spaces. 
We transform all skills to the same label space and give each token a generic tag (i.e., {\texttt{B}, \texttt{I}, \texttt{O}}).
We give a brief description of each dataset below and \cref{tab:num_post} summarizes them:

\paragraph{\textsc{SkillSpan}~\cite{zhang-etal-2022-skillspan}.}
This job posting dataset includes annotations for skills and knowledge derived from the ESCO taxonomy.
To fit our approach, we flatten the two label layers into one layer (i.e., \texttt{BIO}).
The baseline is the JobBERT model, which was continuously pre-trained on a dataset of 3.2 million job posting sentences. 
The industries represented in the data range from tech to more labor-intensive sectors.

\begin{table*}[ht]
\centering
\setlength{\arrayrulewidth}{0.25mm}
\resizebox{0.95\linewidth}{!}{
\begin{tabular}[t]{lllll|l}
\toprule
                                  & \textbf{Setting}                         & \textsc{\textbf{SkillSpan}}        & \textsc{\textbf{Sayfullina}}        &  \textsc{\textbf{Green}}           & \textbf{avg.\ span-F1}    \\\midrule
JobBERT~\cite{zhang-etal-2022-skillspan}  &                                 &  60.47                    &  88.16                     &  42.55                    & 63.73            \\         
\hspace{0.2em} + \knn{}                   & \dwt{}                          &  61.06 \up{0.59}          &  88.25 \up{0.09}           &  43.56 \up{1.01}          & 64.29 \up{0.56}  \\   
\hspace{0.2em} + \knn{}                   & \adwt{}                         &  60.93 \up{0.48}          &  88.26 \up{0.10}           &  44.44 \up{1.89}          & 64.54 \up{0.81} \\\midrule

RoBERTa~\cite{liu2019roberta}             &                                 &  63.88                    &  91.97                     &  44.49                    & 66.78            \\          
\hspace{0.2em} + \knn{}                   & \dwt{}                          &  63.57 \down{0.31}        &  91.97 \none{0.00}         &  45.02    \up{0.53}       & 66.85 \up{0.07}  \\ 
\hspace{0.2em} + \knn{}                   & \adwt{}                         &  63.98 \up{0.10}          &  91.97 \none{0.00}         &  44.86 \up{0.37}          & 66.94 \up{0.16}     \\\midrule    

JobBERTa (This work)                      &                                 &  63.74                    &  92.06                     &  49.61                    & 68.47  \\          
\hspace{0.2em} + \knn{}                   & \dwt{}                          &  64.14 \up{0.40}          &  91.89 \down{0.17}         &  50.35 \up{0.74}          & 68.79 \up{0.32}   \\  
\hspace{0.2em} + \knn{}                   & \adwt{}                         &  \textbf{64.24} \up{0.50}\textsuperscript{$\dagger$}   &  \textbf{92.15} \up{0.09} &  \textbf{50.78} \up{1.17}\textsuperscript{$\dagger$} & \textbf{69.06} \up{0.59}\\          
\bottomrule            
\end{tabular}}
\caption{
\textbf{Test Set Results.} Two settings are considered for each model based on dev.\ set results in~\cref{dev:results}: \dd{} refers to the in-dataset datastore, containing keys from the specific training data, while \ad{} represents a datastore with keys from all available training sets. The notation \texttt{+WT} indicates the application of Whitening Transformation to the keys before adding them to and querying the datastore. The performance impact of using \knn{} is indicated as \up{} (increase), \down{} (decrease), or \textbf{--} (no change). The best-performing setup for each dataset is highlighted. For the top-performing model (JobBERTa), \textsuperscript{$\dagger$} signifies statistical significance over the baseline using a token-level McNemar test~\cite{mcnemar1947note}. The avg.\ span-F1 performance of each model across the three datasets is displayed.
}
\label{tab:res}
\end{table*}

\paragraph{\textsc{Sayfullina}~\cite{sayfullina2018learning}}
is used for soft skill sequence labeling. 
Soft skills are personal qualities that contribute to success, such as teamwork, dynamism, and independence. 
Data originated from the UK.
This is the smallest dataset among the three, with no specified industries.

\paragraph{\textsc{Green}~\cite{green-maynard-lin:2022:LREC}.}
A dataset for extracting skills, qualifications, job domain, experience, and occupation labels. 
The dataset consists of jobs from the UK, and the industries represented include IT, finance, healthcare, and sales. 
This is the largest dataset among the three.

\subsection{Models}

We use 3 English-based LMs: 1 general-purpose and 2 domain-specific models. 
Implementation details for fine-tuning and \nnose{} are in~\cref{sec:params}, including inference costs of our proposed method.

\paragraph{JobBERT~\cite{zhang-etal-2022-skillspan}} is a 110M parameter BERT-based model continuously pre-trained~\cite{gururangan2020don} on 3.2M English job posting sentences. 
It outperforms \bertb{} on several skill-specific tasks.

\paragraph{RoBERTa~\cite{liu2019roberta}.} 
We also use \roberta{} (123M parameters). 
It showed to outperform JobBERT in our initial experiments and we therefore include this model as a baseline.

\paragraph{JobBERTa (Ours).} 
Given that RoBERTa outperformed JobBERT, we create another baseline and release a model named JobBERTa. 
This is a \roberta{} model continuously pre-trained~\cite{gururangan2020don} on the same 3.2M JD sentences as JobBERT.

\section{Results}\label{sec:results}

We evaluate the performance of fine-tuning models enhanced with \nnose{}. 
We consider different setups: First, we compare using the Whitening Transformation (\texttt{+WT}) or without. 
Second, we explore two datastore setups: One using an in-dataset datastore (\dd{}), where each respective training set is stored separately, and another where all datasets are stored in the datastore (\ad{}). 
In the latter setup, we encode all three datasets with each fine-tuned model, and each model has its own \texttt{WT} matrix. 
For example, we fine-tune a model on \textsc{SkillSpan} and encode the training set tokens of \textsc{SkillSpan}, \textsc{Sayfullina}, and \textsc{Green} to populate the datastore. 
From the results on the development set (\cref{tab:dev}, \cref{dev:results}), we observe that adding \texttt{WT} consistently improves performance. 
Therefore, we only report the span-F1 scores on each test set (\cref{tab:res}) \emph{with} \texttt{WT} and the average over all three datasets. 

\paragraph{Best Model Performance.} In~\cref{tab:res}, we show that the best-performing baseline model is JobBERTa, achieving more than 4 points span-F1 improvement over JobBERT and 2 points higher than RoBERTa on average. 
This confirms the effectiveness of DAPT in improving language models~\cite{han-eisenstein-2019-unsupervised,alsentzer2019publicly,gururangan2020don,lee2020biobert,nguyen2020bertweet,zhang-etal-2022-skillspan}.

\paragraph{Best \nnose{} Setting.} 
We confirm the trends from dev.\ on test: The largest improvements come from using the setup with \texttt{WT}, especially in the \adwt{} setting. 
All models seem to benefit from the \nnose{} setup, JobBERT and JobBERTa show the largest improvements, with the largest gains observed in the \adwt{} datastore setup. 
In summary, \adwt{} consistently demonstrates performance enhancements across all experimental setups.

\begin{figure*}[t]
    \centering
    \includegraphics[width=.95\linewidth]{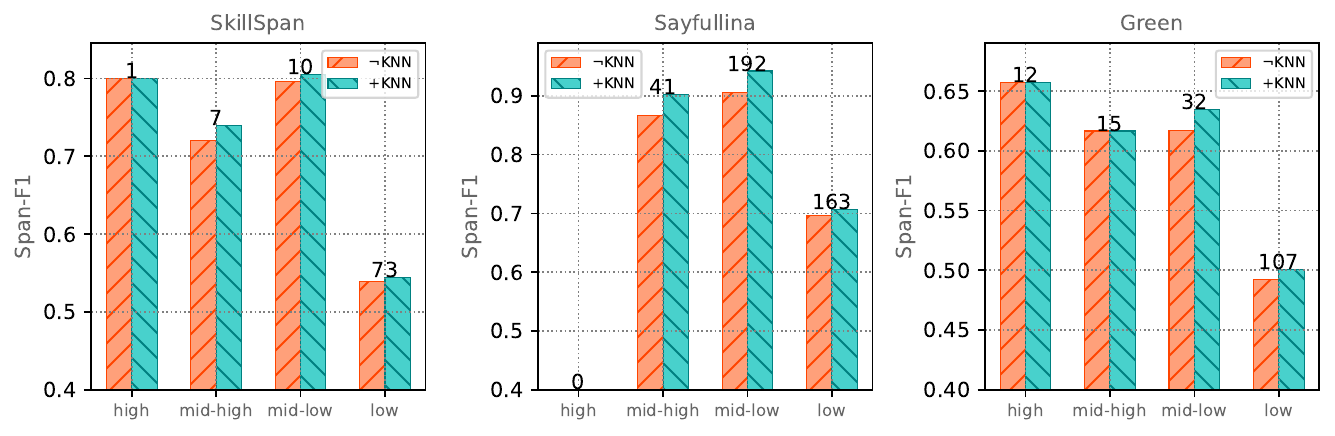}
    \caption{
    \textbf{Long-tail Prediction Performance.} \knn{} is based on the datastore with all the datasets. We categorize the occurrences of a skill in the test set with respect to the training set. For example, a skill in the test set occurs two times in the training set, we put this in the ``low'' bin. There are three frequency ranges: \emph{high}: 10--15, \emph{mid--high}: 7--10, \emph{mid--low}: 4--6, \emph{low}: 0--3. \textsc{Sayfullina} does not have any test set skills that occur more than 10 times in the training set. On top of the bars is the number of predicted skills for the test set in each bucket.
    }
    \label{fig:longtail}
\end{figure*}

\section{Analysis}
As we store training tokens from all datasets in the datastore, we expect the model to recall a greater number of skills based on the current context during inference. In turn, this would lead to improved downstream model performance. We want to address the challenges of SE datasets by predicting long-tail patterns, and if we observe improvements in detecting unseen skills in a cross-dataset setting. 

To investigate in which situations our model improves, we are analyzing the following:
\circled{1} The predictive capability of \nnose{} in relation to rarely occurring skills compared to regular fine-tuning (\cref{subsec:longtail}). Skills exhibit varying frequencies across datasets, we categorize the skill frequencies into buckets and compare the performance between vanilla fine-tuning and the inclusion of \knn{}.
\circled{2} If \nnose{} actually retrieves from other datasets when they are combined (\cref{subsec:all}), and if there is a sign of leveraging multiple datasets, then;
\circled{3} How much does \nnose{} enhance performance in a cross-dataset setting (\cref{subsec:unseen})? Our results indicate a large performance drop when a fine-tuned SE model, trained on one dataset, is applied to another dataset, highlighting the sparsity across datasets. We demonstrate that \nnose{} helps alleviate this, both from an empirical perspective and by inspecting the prediction errors (\cref{subsec:errors}).

\subsection{Long-tail Skills Prediction}\label{subsec:longtail}
\citet{Khandelwal2020Generalization} observed that due to explicitly memorizing the training data, NNLMs effectively predict rare patterns. 
We analyze whether the performance of ``long-tail skills'' improves using \nnose{}. 
A visualization of the long-tail distribution of skills is in \cref{fig:skilldistribution} (\cref{app:skilldistribution}).

We present the results in \cref{fig:longtail}. 
We investigate the performance of JobBERTa with and without \knn{} based on the occurrences of skills in the evaluation set relative to the train set. 
We count the skills in the evaluation set that occur a number of times in the training set, ranging from 0--15 occurrences and is grouped into low, mid--low, mid--high, and high--frequency bins (0--3, 4--6, 7--10, 10--15, respectively). 
This approach estimates the number of skills the LM recalls from the training stage.

Our findings reveal that low-frequent skills are the most difficult and make up the largest bucket, and our approach is able to improve on them on all three datasets.
For \textsc{SkillSpan}, we observe an improvement in the low-frequency bin, from 53.9$\rightarrow$54.5 span-F1. 
Similarly, \textsc{Green} exhibits a similar trend with an improvement in the low-frequency bin (49.2$\rightarrow$50.1). 
Interestingly, it also shows gains in most other frequency bins. 
Last, for \textsc{Sayfullina}, there is also an improvement (69.7$\rightarrow$70.7 in the low bin). 
It is worth pointing out that there are many skills that fall in the low bin in \textsc{SkillSpan} and \textsc{Green}. 
This is exactly where \nnose{} improves most for these datasets. For \textsc{Sayfullina}, we notice the largest number of predicted skills is in the mid--low bin. This is where we also see improvements for \nnose{}.

\subsection{Retrieving From All Datasets}\label{subsec:all}
\begin{figure*}[t]
    \centering
    \includegraphics[width=\linewidth]{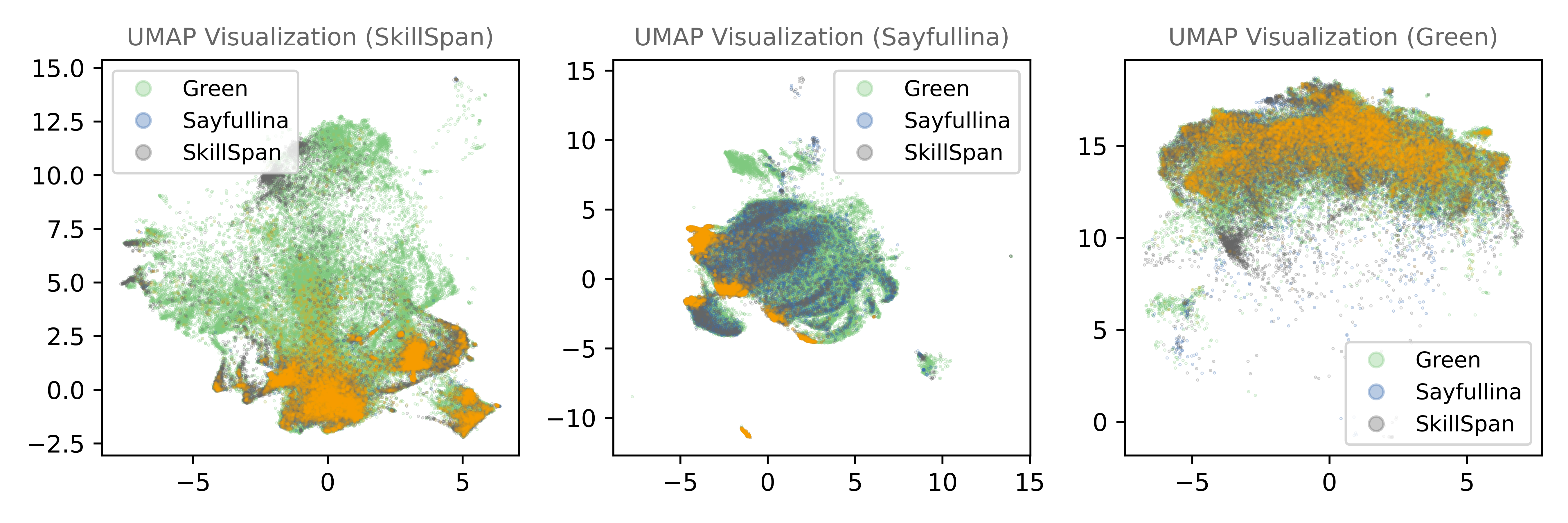}
    \caption{
    \textbf{UMAP Visualization of Nearest Neighbors Retrieval.} The datastore consists of the training set (\texttt{+WT}) of all three datasets used in this work. Each colored dot represents a non-\texttt{O} token from the training set. The embeddings are generated using JobBERTa. The orange shade represents the retrieved neighbors with $k=4$ for each token that is a skill (i.e., not an \texttt{O} token). Note that for the middle plot, the orange shade covers the blue clusters \textsc{Sayfullina}. \textsc{Green} has the green shade and \textsc{SkillSpan} are the darkcyan colors.
    }
    \label{fig:ds-viz}
\end{figure*}

We presented the best improvements of \nnose{} in the \adwt{} datastore in \cref{sec:results}.
An important question remains: Does the \adwt{} setting retrieve from all datasets? 
Qualitatively,~\cref{fig:ds-viz} shows the UMAP visualization \cite{mcinnes2018umap-software} of representations stored in each \adwt{} datastore. 
We mark the retrieved neighbors with orange for each downstream dev.\ set. 
In all plots, we observe that \textsc{Green} is prominent in the representation space (green), while \textsc{SkillSpan} (darkcyan) and \textsc{Sayfullina} (blue) form distinct clusters. 
Each plot has its own pattern: \textsc{SkillSpan} and \textsc{Sayfullina} have well-shaped clusters, while \textsc{Green} consists of one large cluster. 
\textsc{SkillSpan} and \textsc{Sayfullina} mostly retrieve from their own clusters. 
In contrast, \textsc{Green} retrieves from the entire space, which can explain the largest span-F1 performance gains (\cref{tab:res}).
This suggests that \knn{} effectively leverages multiple datasets in most cases. 

\begin{table}[t]
\centering
\setlength{\arrayrulewidth}{0.25mm}
\resizebox{\linewidth}{!}{
\begin{tabular}{lllll}
\toprule
& $\downarrow$Trained on     & \textsc{SkillSpan}        & \textsc{Sayfullina} &  \textsc{Green}      \\\midrule       
\vanilla& \textsc{SkillSpan} &  \cellcolor{gray}         &  18.05              &  43.17               \\          
& \textsc{Sayfullina}        &  9.44                     &  \cellcolor{gray}   &  11.79                \\          
& \textsc{Green}             &  29.67                    &  15.93              &  \cellcolor{gray}    \\\midrule   
& \textsc{All}               &  59.33                    &  90.16              &  44.59               \\\midrule     

\withknn& \textsc{SkillSpan} &  \cellcolor{gray}         &  45.86 \up{27.81}   &  45.44 \up{2.27}     \\          

& \textsc{Sayfullina}        &  26.16 \up{16.72}         &  \cellcolor{gray}   &  25.38 \up{13.59}     \\          
& \textsc{Green}             &  41.22 \up{11.55}         &  46.58 \up{30.65}   &  \cellcolor{gray}     \\\midrule   
& \textsc{All}               &  59.51 \up{0.31}          &  90.33 \up{0.17}    &  45.63 \up{1.04}     \\\bottomrule
\end{tabular}}
\caption{
\textbf{Results of Unseen Skills based on JobBERTa (\adwt{}).} In the vanilla setting, models trained on one skill dataset are applied to another on test, showing varied performance. However, applying \knn{} improves the detection of unseen skills. Diagonal results can be found in~\cref{tab:res}. Refer to~\cref{tab:unseenparams} for tuned hyperparameters.
}
\label{tab:unseen}
\end{table}

\begin{figure*}[t]
    \centering
    \includegraphics[width=.95\linewidth]{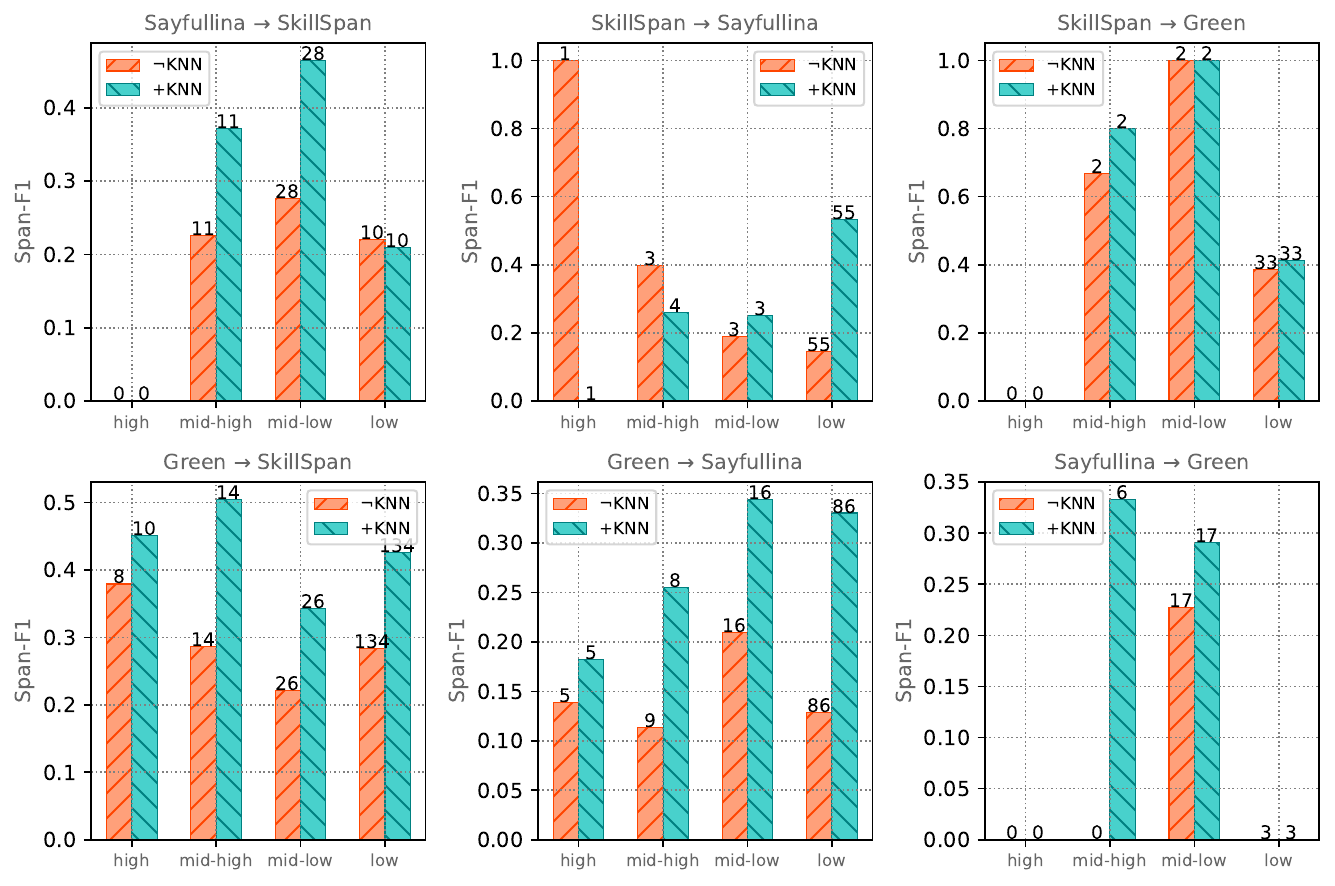}
    \caption{
    \textbf{Cross-dataset Long-tail Performance.} Similar to~\cref{fig:longtail}, we plot the cross-dataset long-tail performance. \nnose{} uses the datastore with all datasets. Training and evaluation data (test) are indicated in graph titles. Frequency bins are based on the training data span frequency; there are three frequency ranges: \emph{high}: 10--15, \emph{mid--high}: 7--10, \emph{mid--low}: 4--6, \emph{low}: 0--3.
    }
    \label{fig:crossdataset}
\end{figure*}

\subsection{Prediction of Unseen Skills}\label{subsec:unseen}
The UMAP plots in \cref{fig:ds-viz} suggest that some datasets are closer to each other than others. 
To quantify this, we investigate the overlap of annotated skills between datasets and assess cross-dataset performance of \nnose{} on unseen skills.

\paragraph{Overlap of Datasets.}
We calculate the exact span overlap of skills between the training sets of the datasets using the Jaccard similarity coefficient \cite{jaccard1901distribution}: $J(A,B) = \frac{|A \cap B|}{|A \cup B|}$, where $A$ and $B$ are sets of multi-token spans (e.g., ``manage a team'') from two separate training sets.
The Jaccard similarity coefficients are as follows: $J$(\textsc{SkillSpan}, \textsc{Sayfullina}) = 0.35, $J$(\textsc{Sayfullina}, \textsc{Green}) = 0.10, and $J$(\textsc{SkillSpan}, \textsc{Green}) = 0.29. 
These Jaccard coefficients indicate overlap between unique skill spans across datasets, suggesting that \nnose{} can introduce the model to new and unseen skills.

\paragraph{Results.}
\cref{tab:unseen} presents the performance of JobBERTa across datasets. 
For completeness, we include a baseline where JobBERTa is fine-tuned on a union of all datasets (\textsc{All}). 
We notice training on the union of the data never leads to the best target dataset performance.
Generally, we observe that in-domain data is best, both in vanilla and NNOSE setups (diagonal in \cref{tab:unseen}). Performance drops when a model is applied to a dataset other than the one it was trained on (off-diagonal). 
Using \nnose{} leads to substantial improvements across the challenging off-diagonal (cross-dataset) settings, while performance remains stable within datasets. 
We observe the largest improvements when applied to \textsc{Sayfullina}, with \emph{up to a 30\%} increase in span-F1. 
This is likely due to \textsc{Sayfullina} consisting mostly of soft skills, which are less prevalent in \textsc{SkillSpan} and \textsc{Green}, making it beneficial to introduce soft skills. 
Conversely, when the model is trained on \textsc{Sayfullina}, the absolute improvement on \textsc{SkillSpan} is lower, indicating that skill datasets can benefit each other to different extents.
 
\paragraph{Cross-dataset Long-tail Analysis.}
\cref{tab:unseen} shows improvements when \nnose{} is used in favor of vanilla fine-tuning. 
\cref{fig:crossdataset} presents the long-tail performance analysis in the cross-dataset scenario, similar to \cref{fig:longtail}. 
We observe the largest gains with \nnose{} in the low or mid--low frequency bins. 
However, exceptions are \textsc{SkillSpan}$\rightarrow$\textsc{Green} and \textsc{Sayfullina}$\rightarrow$\textsc{Green}, where most gains occur in the mid--high bin. 
Notably, \textsc{Sayfullina}$\rightarrow$\textsc{Green} demonstrates higher performance with \nnose{}, where all 6 skills are incorrectly predicted in the mid--high bin.
An analysis of precision and recall in \cref{tab:pr} (\cref{app:cr}) substantiates that the improvements are both precision and recall-based, with gains of up to 40 recall points and 35.4 precision points in \textsc{Green}$\rightarrow$\textsc{Sayfullina}. There is also an improvement up to 35.5 recall points and 34.1 precision points for \textsc{SkillSpan}$\rightarrow$\textsc{Sayfullina}.
This further solidifies that memorizing tokens (i.e., storing all skills in the datastore) helps recall as mentioned in~\citet{Khandelwal2020Generalization}, and more importantly, highlighting the benefits of \nnose{} in cross-dataset scenarios for SE.

\begin{table*}[t]
    \centering
    \resizebox{.95\linewidth}{!}{
    \begin{tabular}{lll}
    \toprule
                           & \textbf{False Positives}                           & \textbf{False Negatives}  \\\midrule
                           & cleaning                                           & GCP \\
    \textsc{SkillSpan}     & decisive                                           & IBM MQ \\
                           & Apache Camel                                       & AWS \\
                           & building consumer demand for sustainable products  & budget responsible\\\midrule
                           & empathy                                            & leadership      \\
      \textsc{Sayfullina}  & leadership management                              &                 \\
                           & communication                                      &                 \\
                           & ability to manage and prioritise multiple assignments and tasks &    \\\midrule
                           & SQL scripting languages                            & software engineering \\
          \textsc{Green}   & Manage a team                                      & development \\
                           & troubleshooting activities                         & DevOps \\
                           & dealing with tenants                               & Cisco network administration \\\bottomrule
    \end{tabular}
    }
    \caption{\textbf{FPs \& FNs of \nnose{}.} We show several examples of false positives and false negatives in each dataset. We only show the predictions of \nnose{} that are \emph{not} in the vanilla model predictions.}
    \label{tab:fpfn}
\end{table*}

\subsection{Qualitative Check on Prediction Errors}\label{subsec:errors}
We perform a qualitative analysis on the false positives (fp) and false negatives (fn) of \nnose{} predictions compared to vanilla fine-tuning for each dataset. 
This analysis tells us whether a prediction corresponds to an actual skill, even if it does not contribute positively to the span-F1 metric. 
We observe that \nnose{} 
produces a significant number of false positives that are ``similar'' to genuine skills. 
In~\cref{tab:fpfn}, for each dataset, we picked five fps and fns that represent hard, soft, and personal skills well (if applicable). We show the fps and fns for JobBERTa with \nnose{}, we only show predictions that are \emph{not} in the vainlla model predictions. 
In \textsc{Sayfullina}, there is only one fn. We notice from the errors, and especially the fps, that these are definitely skills, indicating the benefit of \nnose{} helping to predict new skills or missed annotations. For a general qualitative check on predictions, we refer to~\cref{app:qualitative}. We show that \nnose{} predicts a variety of close tokens, but also the same tokens if the model is confident about the predictions (i.e., high softmax scores).

\section{Related Work}
\paragraph{Skill Extraction.}
The dynamic nature of labor markets has led to an increase in tasks related to JD, including skill extraction~\cite{kivimaki-etal-2013-graph,zhao2015skill,sayfullina2018learning,smith2019syntax,tamburri2020dataops,shi2020salience, chernova2020occupational,bhola-etal-2020-retrieving,gugnani2020implicit,fareri2021skillner,konstantinidis2022knowledge, zhang-etal-2022-skillspan,zhang-jensen-plank:2022:LREC,zhang2022skill,green-maynard-lin:2022:LREC,gnehm-bhlmann-clematide:2022:LREC,beauchemin2022fijo,decorte2022design, ao2023skill, goyal-etal-2023-jobxmlc, zhang2023escoxlmr}. 
These works employ methods such as sequence labeling \cite{sayfullina2018learning, smith2019syntax, chernova2020occupational, zhang-etal-2022-skillspan, zhang2022skill}, multi-label classification \cite{bhola-etal-2020-retrieving}, and graph-based methods \cite{shi2020salience, goyal-etal-2023-jobxmlc}. 
Recent methodologies include domain-specific models where LMs are continuously pre-trained on unlabeled JD \cite{zhang-etal-2022-skillspan, gnehm-bhlmann-clematide:2022:LREC}. 
However, none of these methodologies introduce a retrieval-augmented model like \nnose{}.

\paragraph{General Retrieval-augmentation.}
In retrieval augmentation, LMs can utilize external modules to enhance their context-processing ability.
Two approaches are commonly used: First, using a separately trained model to retrieve relevant documents from a collection. 
This approach is employed in open-domain question answering tasks~\cite{petroni-etal-2021-kilt} and with specific models such as ORQA~\cite{lee-etal-2019-latent}, REALM~\cite{guu2020retrieval}, RAG~\cite{lewis2020retrieval}, FiD~\cite{izacard2021distilling}, and ATLAS~\cite{izacard2022few}.

Second, previous work on explicit memorization showed promising results with a cache~\cite{grave2017improving}, which serves as a type of datastore. 
The cache contains past hidden states of the model as keys and the next word as tokens in key--value pairs. Memorization of hidden states in a datastore, involves using the \knn{} algorithm as the retriever. 
The first work of the \knn{} algorithm as the retrieval component was by \citet{Khandelwal2020Generalization}, leading to several LM decoder-based works.

\paragraph{Decoder-based Nearest Neighbor Approaches.}
Decoder-based nearest neighbors approaches are primarily focused on language modeling~\cite{Khandelwal2020Generalization, he-etal-2021-efficient, yogatama2021adaptive, ton2022regularized, shi2022nearest, jin2022plug, bhardwaj2022adaptation, xu2023nearest} and machine translation~\cite{khandelwal2021nearest, zheng2021adaptive, jiang2021learning, jiang2022towards, wang2022efficient, martins2022efficient, martins2022chunk, zhu2022knowledge, du2023federated, zhu2023knn, min-etal-2023-nonparametric, min2023silo}. 
These approaches often prioritize efficiency and storage space reduction, as the datastores for these tasks can contain billions of tokens.

\paragraph{Encoder-based Nearest Neighbor Approaches.}
Encoder-based nearest neighbor approaches have been explored in tasks such as named entity recognition~\cite{wang2022k} and emotion classification~\cite{yin2022efficient}. 
Here, the datastores are limited to single datasets with the sentence (or token) gold label pairs. 
Instead, we show the potential of adding multiple datasets in the datastore.

\section{Conclusion}
We introduce \nnose{}, an LM that incorporates and leverages a non-parametric datastore for nearest neighbor retrieval of skill tokens. 
To the best of our knowledge, we are the first to introduce the nearest neighbors retrieval component for the extraction of occupational skills. 
We evaluated \nnose{} on three relevant skill datasets with a wide range of skills and show that \nnose{} enhances the performance of all LMs used in this work \emph{without} additionally tuning the LM parameters. 
Through the combination of train sets in the datastore, our analysis reveals that \nnose{} effectively leverages all the datasets by retrieving from each. 
Moreover, \nnose{} not only performs well on rare skills but also enhances the performance on more frequent patterns. 
Lastly, we observe that our baseline models exhibit poor performance when applied in a cross-dataset setting. 
However, with the introduction of \nnose{}, the models improve across all settings. 
Overall, our findings indicate that \nnose{} is a promising approach for application-specific skill extraction setups and potentially helps discover skills that were missed in manual annotations.

\section*{Limitations}
We consider several limitations: One is the limited diversity of the datasets used in this work. 
Our study was constrained by the use of only three English datasets. 
By focusing solely on English data, the method might not generalize other languages. 

Future research includes incorporating a wider range of datasets from diverse sources to obtain a more comprehensive understanding of the topic. 
Potential interesting future work should include validation on whether \nnose{} works in a multilingual setting.

Another limitation is that we do skill detection and not specific labeling of the extracted spans, i.e., extracting generic \texttt{B}, \texttt{I}, \texttt{O} tags.
This was to ensure that the datasets could be used all together in the datastore. 
%Interesting future work could be to extend \nnose{} to include labeled skills in the datastore.

Last, we only applied the nearest neighbors with the datastore to the job market domain. In contrast,~\citet{wang2022k} have used a similar approach on  a more generic domain, e.g, CoNLL data~\citep{tjong-kim-sang-de-meulder-2003-introduction}, but also keep it limited to the number of labels in this dataset (i.e., four fine-grained labels: \texttt{Person}, \texttt{Location}, \texttt{Organization}, and \texttt{Misc.}). We believe with coarse-grained span labeling (i.e., \texttt{BIO}), our proposed method and positive results have the potential to transfer to other domains.

\section*{Ethics Statement}

The subject of job-related language models is a highly contentious topic, often sparking intense debates surrounding the issue of bias. 
We acknowledge that LMs such as JobBERTa and \nnose{} possess the potential for inadvertent consequences, such as unconscious bias and dual-use when employed in the candidate selection process for specific job positions. 
There are research efforts to develop fairer recommender systems in the field of human resources, focusing on mitigating biases (e.g., \citealp{mujtaba2019ethical, raghavan2020mitigating, deshpande2020mitigating, kochling2020discriminated, sanchez2020does, wilson2021building, vanimproving, arafan2022end}). 
Nevertheless, one potential approach to alleviating such biases involves the retrieval of sparse skills for recall (e.g., this work). 
It is important to note, however, that we have not conducted an analysis to ascertain whether this particular method exacerbates any pre-existing forms of bias.

\section*{Acknowledgements}
We thank the MaiNLP and NLPnorth group for feedback on an earlier version of this paper, and WING for hosting MZ for a research stay. In particular, thanks to Elisa Bassignana, Robert Litschko, Max M{\"u}ller-Eberstein, Yanxia Qin, and Tongyao Zhu for helpful suggestions and feedback. This research is supported by the Independent Research Fund Denmark (DFF) grant 9131-00019B and in parts by ERC Consolidator Grant DIALECT 101043235. 

\bibliography{anthology,custom}

\clearpage
\appendix

\begin{figure*}[!htb]
\begin{minipage}{0.3\textwidth}
\begin{minted}[frame=single,
               framesep=3mm,
               linenos=true,
               xleftmargin=15pt,
               tabsize=2]{xml}
Experience	O
in	        O
working     B
on          I
a           I
cloud-based I
application I
running     O
on          O
Docker      B
.           O

A           O
degree      B
in          I
Computer    I
Science     I
or          O
related     O
fields      O
.           O
\end{minted}
\caption{\textbf{Data Example for SkillSpan.} In \textsc{SkillSpan}, note the long skills.} 
\label{json-skillspan}
\end{minipage}\hfill
\begin{minipage}{0.3\textwidth}
\begin{minted}[frame=single,
               framesep=3mm,
               linenos=true,
               xleftmargin=15pt,
               tabsize=2]{xml}
ability     O
to          O
work        B
under       I
stress      I
condition   O

due         O
to          O
the         O
dynamic     B
nature      O
of          O
the         O
group       O
environment O
,           O
the         O
ideal       O
candidate   O
will        O
\end{minted}
\caption{\textbf{Data Example for Sayfullina.} In \textsc{Sayfullina}, the skills are usually soft-like skills.} 
\label{json-sayfullina}
\end{minipage}\hfill
\begin{minipage}{0.3\textwidth}
\begin{minted}[frame=single,
               framesep=3mm,
               linenos=true,
               xleftmargin=15pt,
               tabsize=2]{xml}
A               O
sound           O
understanding   O
of              O
the             O
Care            B
Standards       I
together        O
with            O
a               O
Nursing         B
qualification   I
and             O
current         O
NMC             B
registration    I
are             O
essential       O
for             O
this            O
role            O
\end{minted}
\caption{\textbf{Data Example for Green.} There are many qualification skills (e.g., certificates).} 
\label{json-green}
\end{minipage}
\end{figure*}

\section{Whitening Transformation Algorithm}\label{app:wt}

\begin{algorithm}[htbp]
\SetAlgoLined
 \textbf{input:} Embeddings $\left\{x_i\right\}_{i=1}^N$;\\

 Compute $\mu=\frac{1}{N} \sum_{i=1}^N x_i$ and $\Sigma$ of $\left\{x_i\right\}_{i=1}^N$

 Compute $U, \Lambda, U^\top = \text{SVD}(\Sigma)$

 Compute $W = U\sqrt{\Lambda^{-1}}$

 \For{$i = 1, 2, ..., n$}{
    $\widetilde{x}_i=\left(x_i-\mu\right) W$
 
 }
 \textbf{return} $\left\{\widetilde{x}_i\right\}_{i=1}^N$;
 \caption{Whitening Transformation Workflow}
 \label{algo:white}
\end{algorithm}

We apply the whitening transformation to the query embedding and the embeddings in the datastore. 
We can write a set of token embeddings as a set of row vectors: $\left\{x_i\right\}_{i=1}^N$. Additionally, a linear transformation $\widetilde{x}_i=\left(x_i-\mu\right) W$ is applied, where $\mu=\frac{1}{N} \sum_{i=1}^N x_i$. 
To obtain the matrix $W$, the following steps are conducted: First, we obtain the original covariance matrix

\begin{equation}
    \Sigma=\frac{1}{N} \sum_{i=1}^N\left(x_i-\mu\right)^\top\left(x_i-\mu\right).
\end{equation}

Afterwards, we obtain the transformed covariance matrix $\widetilde{\Sigma}=W^\top \Sigma W$, where we specify $\widetilde{\Sigma}=I$. 
Therefore, $\Sigma =\left(W^\top\right)^{-1} W^{-1} = \left(W^{-1}\right)^\top W^{-1}$. 
Here, $\Sigma$ is a positive definite symmetric matrix that satisfies the following singular value decomposition (SVD;~\citealp{golub1971singular}) as indicated by \citet{su2021whitening}:
$\Sigma= U\Lambda U^\top.$
$U$ is an orthogonal matrix, $\Lambda$ is a diagonal matrix, and the diagonal elements are all positive. 
Therefore, let $W^{-1}=\sqrt{\Lambda}U^\top$, we obtain the solution: $W = U\sqrt{\Lambda^{-1}}$. Putting it all together, as input, we have the set of embeddings $\left\{x_i\right\}_{i=1}^N$. We compute $\mu$ and $\Sigma$ of $\left\{x_i\right\}_{i=1}^N$. 
Then, we perform SVD on $\Sigma$ to obtain matrices $U$, $\Lambda$, and $U^\top$. Using these matrices, we calculate the transformation matrix $W$. 
Finally, we apply the transformation to each embedding in the set by subtracting $\mu$ and multiplying by $W$. We are left with $\widetilde{x}_i=\left(x_i-\mu\right) W$. Note that we do \texttt{WT} \emph{before} we store the embedding in the datastore, and apply \texttt{WT} to the token embedding before we query the datastore.

We show the Whitening Transformation procedure in~\cref{algo:white}. Note that \citet{li-etal-2020-sentence, su2021whitening} introduced a dimensionality reduction factor $k$ on $W$ ($W[:, :k]$). 
The diagonal elements in the matrix $\Lambda$ obtained from the SVD algorithm are in descending order. 
One can decide to keep the first $k$ columns of $W$ in line 6. 
This is similar to PCA \citep{abdi2010principal}. 
However, empirically, we found that reducing dimensionality had a negative effect on downstream performance, thus we omit that in this implementation.

\section{Data Examples}\label{data:examples}

\begin{table}[ht]
    \centering
    \begin{tabular}{ll}
    \toprule
    \textsc{SkillSpan}      & \cref{json-skillspan} \\
    \textsc{Sayfullina}     & \cref{json-sayfullina} \\
    \textsc{Green}          & \cref{json-green} \\
    \bottomrule
    \end{tabular}
    \caption{Data example references for each dataset.}
    \label{tab:dataexamples}
\end{table}

In \cref{tab:dataexamples}, we refer to several listings of examples of the datasets. Notably in \textsc{SkillSpan}, the original samples contain two columns of labels. These refer to skills and knowledge. To accommodate for the approach of \nnose{}, we merge the labels together and thus removing the possible nesting of skills.~\citet{zhang-etal-2022-skillspan} mentions that there is not a lot of nesting of skills. Following~\citet{zhang-etal-2022-skillspan}, we prioritize the skills column when merging the labels. When there is nesting, we keep the labels of skills and remove the knowledge labels.

\begin{table*}[!ht]
\begin{minipage}{0.48\textwidth}
\centering
\resizebox{\linewidth}{!}{
\begin{tabular}[t]{rrlll}
\toprule
 Dataset $\rightarrow$          &                                    & \textsc{SkillSpan}   & \textsc{Sayfullina} &  \textsc{Green}                       \\\midrule
JobBERT                         & $k$                                & 4                    & 4                   &  16                   \\
                                & $\lambda$                          & 0.3                  & 0.3                 &  0.15                 \\
                                & $T$                                & 0.1                  & 2.0                 &  10.0                 \\\midrule
RoBERTa                         & $k$                                & 32                   & 4                   & 64                    \\
                                & $\lambda$                          & 0.3                  & 0.3                 & 0.25                  \\
                                & $T$                                & 10.0                 & 0.1                 & 10.0                  \\\midrule
JobBERTa                        &  $k$                               & 16                   & 4                   & 8                     \\
                                &  $\lambda$                         & 0.2                  & 0.1                 & 0.1                   \\
                                &  $T$                               & 5.0                  & 10.0                & 10.0                  \\
\midrule
                                &  $k$                               & \multicolumn{3}{c}{\{4, 8, 16, 32, 64, 128\}}                     \\
Search Space                    &  $\lambda$                         & \multicolumn{3}{c}{\{0.1, 0.15, 0.2, 0.25, ..., 0.9\}}            \\
                                &  $T$                               & \multicolumn{3}{c}{\{0.1, 0.5, 1.0, 2.0, 3.0, 5.0, 10.0\}}       \\\bottomrule
\end{tabular}}
\caption{\textbf{Tuned Hyperparameters on Dev.} These are for $\{\mathcal{D}\}$.}
\label{tab:params1}
\end{minipage}\hfill
\begin{minipage}{0.48\textwidth}
\centering
\resizebox{\linewidth}{!}{
\begin{tabular}[t]{rrlll}
\toprule
 Dataset $\rightarrow$          &                                    & \textsc{SkillSpan}   & \textsc{Sayfullina} &  \textsc{Green}                       \\\midrule
JobBERT                         & $k$                                & 4                    & 4                   &  64                   \\
                                & $\lambda$                          & 0.35                 & 0.35                &  0.4                  \\
                                & $T$                                & 2.0                  & 0.1                 &  5.0                  \\\midrule
RoBERTa                         & $k$                                & 32                   & 4                   & 16                    \\
                                & $\lambda$                          & 0.35                 & 0.45                & 0.25                  \\
                                & $T$                                & 0.1                  & 0.1                 & 1.0                   \\\midrule
JobBERTa                        &  $k$                               & 64                   & 128                 & 128                   \\
                                &  $\lambda$                         & 0.25                 & 0.35                & 0.45                   \\
                                &  $T$                               & 10.0                 & 0.5                 & 10.0                  \\
\midrule
                                &  $k$                               & \multicolumn{3}{c}{\{4, 8, 16, 32, 64, 128\}}                     \\
Search Space                    &  $\lambda$                         & \multicolumn{3}{c}{\{0.1, 0.15, 0.2, 0.25, ..., 0.9\}}            \\
                                &  $T$                               & \multicolumn{3}{c}{\{0.1, 0.5, 1.0, 2.0, 3.0, 5.0, 10.0\}}       \\\bottomrule
\end{tabular}}
\caption{\textbf{Tuned Hyperparameters on Dev.} These are for $\{\mathcal{D}\} +WT$.}
\label{tab:params2}
\end{minipage}\hfill
\begin{minipage}{0.48\textwidth}
\centering
\resizebox{\linewidth}{!}{
\begin{tabular}[t]{rrlll}
\toprule
 Dataset $\rightarrow$          &                                    & \textsc{SkillSpan}   & \textsc{Sayfullina} &  \textsc{Green}                       \\\midrule
JobBERT                         & $k$                                & 4                    & 16                  &  32                   \\
                                & $\lambda$                          & 0.3                  & 0.25                &  0.15                 \\
                                & $T$                                & 10.0                 & 5.0                 &  10.0                 \\\midrule
RoBERTa                         & $k$                                & 16                   & 8                   & 8                     \\
                                & $\lambda$                          & 0.15                 & 0.1                 & 0.1                   \\
                                & $T$                                & 10.0                 & 10.0                & 10.0                  \\\midrule
JobBERTa                        &  $k$                               & 8                    & 4                   & 8                     \\
                                &  $\lambda$                         & 0.2                  & 0.15                & 0.1                   \\
                                &  $T$                               & 0.5                  & 0.1                 & 10.0                  \\
\midrule
                                &  $k$                               & \multicolumn{3}{c}{\{4, 8, 16, 32, 64, 128\}}                     \\
Search Space                    &  $\lambda$                         & \multicolumn{3}{c}{\{0.1, 0.15, 0.2, 0.25, ..., 0.9\}}            \\
                                &  $T$                               & \multicolumn{3}{c}{\{0.1, 0.5, 1.0, 2.0, 3.0, 5.0, 10.0\}}       \\\bottomrule
\end{tabular}}
\caption{\textbf{Tuned Hyperparameters on Dev.} These are for \ad{}.}
\label{tab:params3}
\end{minipage}\hfill
\begin{minipage}{0.48\textwidth}
\centering
\resizebox{\linewidth}{!}{
\begin{tabular}[t]{rrlll}
\toprule
 Dataset $\rightarrow$          &                                    & \textsc{SkillSpan}   & \textsc{Sayfullina} &  \textsc{Green}                       \\\midrule
JobBERT                         & $k$                                & 32                   & 4                   &  128                  \\
                                & $\lambda$                          & 0.3                  & 0.3                 &  0.4                  \\
                                & $T$                                & 1.0                  & 0.5                 &  2.0                  \\\midrule
RoBERTa                         & $k$                                & 128                  & 128                 & 64                    \\
                                & $\lambda$                          & 0.35                 & 0.1                 & 0.25                   \\
                                & $T$                                & 0.1                  & 0.5                 & 0.1                   \\\midrule
JobBERTa                        &  $k$                               & 32                   & 8                   & 128                   \\
                                &  $\lambda$                         & 0.15                 & 0.3                 & 0.2                   \\
                                &  $T$                               & 0.1                  & 0.1                 & 2.0                   \\
\midrule
                                &  $k$                               & \multicolumn{3}{c}{\{4, 8, 16, 32, 64, 128\}}                     \\
Search Space                    &  $\lambda$                         & \multicolumn{3}{c}{\{0.1, 0.15, 0.2, 0.25, ..., 0.9\}}            \\
                                &  $T$                               & \multicolumn{3}{c}{\{0.1, 0.5, 1.0, 2.0, 3.0, 5.0, 10.0)\}}       \\\bottomrule
\end{tabular}}
\caption{\textbf{Tuned Hyperparameters on Dev.} These are for \adwt{}.}
\label{tab:params4}
\end{minipage}\hfill
\end{table*}

\begin{table}
    \centering
    \resizebox{\linewidth}{!}{
\begin{tabular}{lllll}
\toprule
             $\downarrow$Trained on                & Hyperparams.\                      & \textsc{SkillSpan}   & \textsc{Sayfullina} &  \textsc{Green}        \\\midrule          
             \textsc{SkillSpan}         &  $k$                               & \cellcolor{gray!25}  & 16                  & 32                   \\
                                        &  $\lambda$                         & \cellcolor{gray!25}  & 0.9                 & 0.7                   \\
                                        &  $T$                               & \cellcolor{gray!25}  & 0.1                 & 0.5                   \\\midrule 
             \textsc{Sayfullina}        &  $k$                               & 64                   & \cellcolor{gray!25} & 32                   \\
                                        &  $\lambda$                         & 0.9                  & \cellcolor{gray!25} & 0.8                   \\
                                        &  $T$                               & 0.1                  & \cellcolor{gray!25} & 0.1                   \\\midrule
             \textsc{Green}             &  $k$                               & 32                   & 32                  & \cellcolor{gray!25}   \\
                                        &  $\lambda$                         & 0.85                 & 0.9                 & \cellcolor{gray!25}   \\
                                        &  $T$                               & 0.5                  & 0.1                 & \cellcolor{gray!25}   \\\midrule
             \textsc{All}               &  $k$                               & 4                    & 128                 & 32                  \\
                                        &  $\lambda$                         & 0.25                 & 0.6                 & 0.65                  \\
                                        &  $T$                               & 1.0                  & 1.0                 & 0.5                  \\
                                        \midrule
                                        &  $k$                               & \multicolumn{3}{c}{\{4, 8, 16, 32, 64, 128\}}                     \\
             Search Space               &  $\lambda$                         & \multicolumn{3}{c}{\{0.1, 0.15, 0.2, 0.25, ..., 0.9\}}            \\
                                        &  $T$                               & \multicolumn{3}{c}{\{0.1, 0.5, 1.0, 2.0, 3.0, 5.0, 10.0\}}       \\\bottomrule
\end{tabular}}
\caption{
\textbf{Results of Unseen Skills (Development Set) based on JobBERTa.}
}
\label{tab:unseenparams}
\end{table}

\section{Implementation Details}
\label{sec:params}

\paragraph{General Implementation.} We obtain all LMs from the Transformers library~\cite{wolf2020transformers} and implement JobBERTa using the same library. All learning rates for fine-tuning are $5 \times 10^{-5}$ using the AdamW optimizer~\cite{loshchilov2017decoupled}. We use a batch size of 16 and a maximum sequence length of 128 with dynamic padding. The models are trained for 20 epochs with early stopping using a patience of 5. We implement the retrieval component using the FAISS library~\cite{johnson2019billion}, which is a standard for nearest neighbors retrieval-augmented methods.\footnote{\url{https://faiss.ai/}} %For further details and hyperparameter-specific settings for \nnose{}, we refer to~\cref{sec:params}.

\paragraph{JobBERTa.} We apply domain-adaptive pre-training~\cite{gururangan2020don}, which involves continued self-supervised pre-training of a large LM on domain-specific text. This approach enhances the modeling of text for downstream tasks within the domain. We continue pre-training on a \texttt{roberta-base} checkpoint with 3.2M job posting sentences from~\citet{zhang-etal-2022-skillspan}. We use a batch size of 8 and run MLM for a single epoch following~\citet{gururangan2020don}. The rest of the hyperparameters are set to the defaults in the Transformer library.\footnote{\url{https://github.com/huggingface/transformers/blob/main/examples/pytorch/language-modeling/run_mlm.py}}

\paragraph{\nnose{} Setup.}
Following previous work, the keys used in \nnose{} are the 768-dimensional representation logits obtained from the final layer of the LM (input to the softmax). We perform a single forward pass over the training set of each dataset to save the keys and values, i.e., the hidden representation and the corresponding gold BIO tag. The FAISS index is created using all the keys to learn 4096 cluster centroids. During inference, we retrieve $k$ neighbors. The index looks up 32 cluster centroids while searching for the nearest neighbors. For all experiments, we compute the squared Euclidean ($L^2$) distances with full precision keys. The difference in inference speed is almost negligible, with the \knn{} module taking a few extra seconds compared to regular inference. For the exact hyperparameter values, we indicate them in the next paragraph.

\begin{table*}[t]
\centering
\resizebox{\linewidth}{!}{
\begin{tabular}{llllll}%lll|}
\toprule
Dataset (Dev.) $\rightarrow$                 & Setting                                 & \textsc{SkillSpan}   & \textsc{Sayfullina}    &  \textsc{Green}           & avg.\ Span-F1\\\midrule          
JobBERT~\cite{zhang-etal-2022-skillspan}     &                                         &  61.08               &  89.26                 &  37.27                    & 62.54           \\         
\hspace{0.2em} + \knn{}                      & \dd{}                                   &  61.56 \up{0.48}     &  89.69 \up{0.43}       &  37.48 \up{0.21}          & 62.91 \up{0.37} \\ 
\hspace{0.2em} + \knn{}                      & \dwt{}                                  &  61.77 \up{0.69}     &  89.78 \up{0.52}       &  38.07 \up{0.80}          & 63.21 \up{0.67} \\   
\hspace{0.2em} + \knn{}                      & \ad{}                                   &  61.58 \up{0.50}     &  89.50 \up{0.24}       &  37.27 \none{0.00}        & 62.78 \up{0.24} \\
\hspace{0.2em} + \knn{}                      & \adwt{}                                 &  61.50 \up{0.42}     &  89.37 \up{0.11}       &  38.19 \up{0.92}          & 63.02 \up{0.48} \\\midrule 

RoBERTa~\cite{liu2019roberta}                &                                         &  65.02               &  92.91                 &  40.33                    & 66.09  \\          
\hspace{0.2em} + \knn{}                      & \dd{}                                   &  65.36 \up{0.34}     &  92.76 \down{0.15}     &  40.53 \up{0.20}          & 66.22 \up{0.13} \\ 
\hspace{0.2em} + \knn{}                      & \dwt{}                                  &  65.34 \up{0.32}     &  93.07 \up{0.16}       &  41.22 \up{0.89}          & 66.54 \up{0.45} \\         
\hspace{0.2em} + \knn{}                      & \ad{}                                   &  64.98 \down{0.04}   &  92.78 \down{0.13}     &  40.60 \up{0.27}          & 66.12 \up{0.03} \\  
\hspace{0.2em} + \knn{}                      & \adwt{}                                 &  65.38 \up{0.36}     &  92.92 \up{0.01}       &  41.11 \up{0.77}          & 66.47 \up{0.38} \\\midrule    

JobBERTa (This work)                         &                                         &  65.15               &  92.09                 &  40.59                    & 65.94  \\          
\hspace{0.2em} + \knn{}                      & \dd{}                                   &  65.25 \up{0.10}     &  91.99 \down{0.10}     &  41.31 \up{0.72}          & 66.18 \up{0.24} \\ 
\hspace{0.2em} + \knn{}                      & \dwt{}                                  &  65.21 \up{0.06}     &  92.10 \up{0.01}       &  41.41 \up{0.82}          & 66.24 \up{0.30} \\        
\hspace{0.2em} + \knn{}                      & \ad{}                                   &  65.15 \none{0.00}   &  92.04 \down{0.05}     &  40.83 \up{0.24}          & 66.01 \up{0.07} \\   
\hspace{0.2em} + \knn{}                      & \adwt{}                                 &  65.22 \up{0.07}     &  92.13 \up{0.04}       &  41.45 \up{0.86}          & 66.26 \up{0.32} \\          
\bottomrule              
\end{tabular}}
\caption{\textbf{Development Set Results.} There are four settings for each model. \dd{}: in-dataset datastore (i.e., the datastore only contains the keys from the specific training data it is applied on). \ad{}: The datastore contains the keys from all available training datasets. $+W$: Whitening Transformation is applied to the keys before adding them to the datastore or querying the datastore. We indicate the performance increase (\up{}), decrease (\down{}), or no change (\textbf{--}) when using \knn{} compared to not using \knn{}. Additionally, we show the average span-F1 performance of each model across the three datasets. In the development set, it seems that an in-dataset datastore works best.}
\label{tab:dev}
\end{table*}

\paragraph{Hyperparameters \nnose{}.} The best-performing hyperparameters and search space can be found in~\cref{tab:params1}, \cref{tab:params2}, \cref{tab:params3}, and \cref{tab:params4}. We report the $k$-nearest neighbors, $\lambda$ value, and softmax temperature $T$ for each dataset and model.

In \cref{tab:unseenparams}, we show the hyperparameters for the cross-dataset analysis. In the vanilla setting, we apply the models trained on a particular skill dataset to another skill dataset, similar to transfer learning. We observe a significant discrepancy in performances cross-dataset, indicating a wide range of skills. However, when \knn{} is applied, it improves the detection of unseen skills. The datastore contains tokens from all datasets.

\begin{figure*}[ht]
    \centering
    \includegraphics[width=.85\linewidth]{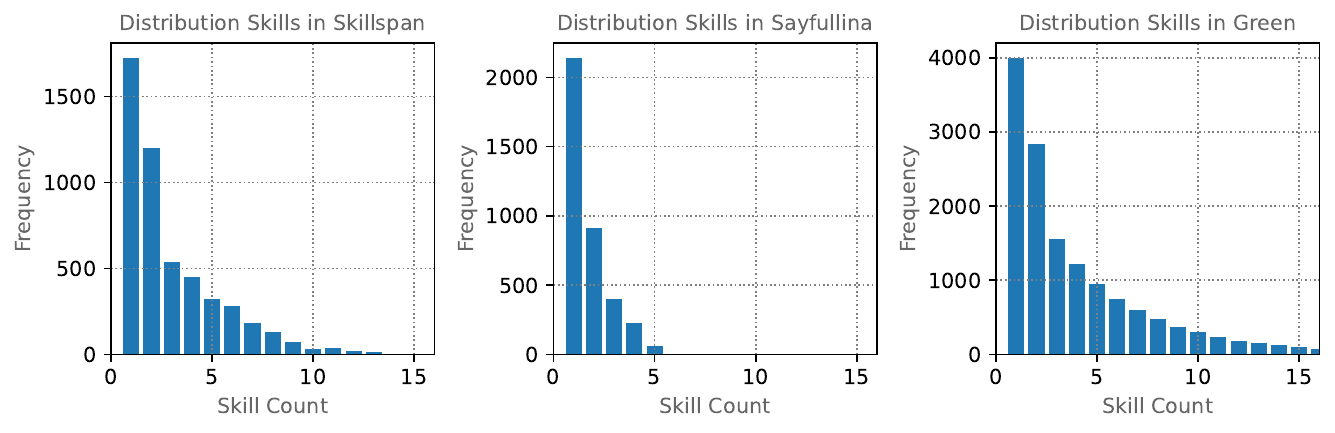}
    \caption{
    \textbf{Frequency Distribution of Skill Occurrences in the Train Set.} We display the frequency distribution of skill occurrences in each train set. \emph{How to read}: For instance, in the case of Sayfullina, there are over 2,000 skills that occur only \textbf{once} in the training set. We demonstrate that all skill datasets exhibit an inherent long-tail pattern.
    }
    \label{fig:skilldistribution}
\end{figure*}

\begin{table*}[t]
    \centering
    \resizebox{.65\linewidth}{!}{
    \begin{tabular}{lllll}
    \toprule
                                                           & \multicolumn{2}{c}{Vanilla}                   & \multicolumn{2}{c}{+\knn{}} \\
    Setup$\downarrow$                                      & Precision            & Recall                  & Precision             & Recall    \\\midrule
    \textsc{Sayfullina}$\rightarrow$\textsc{SkillSpan}     & 10.20                & 10.50                   & 37.67\up{27.47}       & 29.62\up{19.12}     \\
    \textsc{Green}$\rightarrow$\textsc{SkillSpan}          & 28.40                & 33.56                   & 46.00\up{11.60}       & 46.29\up{12.73}     \\\midrule
    \textsc{SkillSpan}$\rightarrow$\textsc{Sayfullina}     & 15.19                & 23.42                   & 49.25\up{34.06}       & 58.95\up{35.53}     \\
    \textsc{Green}$\rightarrow$\textsc{Sayfullina}         & 12.80                & 21.58                   & 48.21\up{35.41}       & 61.87\up{40.29}     \\\midrule
    \textsc{SkillSpan}$\rightarrow$\textsc{Green}          & 52.01                & 37.42                   & 55.37\up{3.36}        & 38.74\up{1.32}     \\
    \textsc{Sayfullina}$\rightarrow$\textsc{Green}         & 17.79                & 7.64                    & 39.83\up{22.04}       & 18.31\up{10.67}     \\\bottomrule
    \end{tabular}
    }
    \caption{\textbf{Precision \& Recall Numbers Cross-dataset on Test.} We show the precision and recall numbers in the cross-dataset setup. We use the \adwt{} setup here, with JobBERTa as the backbone model.}
    \label{tab:pr}
\end{table*}

\paragraph{Inference Cost.}
Due to the current size of the datasets (less than 1M tokens in total), it has no noticeable effect on inference time with the fast nearest neighbor search of FAISS~\citep{johnson2019billion}. We imagine if the datasets come closer to billions of tokens e.g., in machine translation~\citep{khandelwal2021nearest} and language modeling~\citep{Khandelwal2020Generalization}, the inference time will be larger.

\section{Development Set Results}\label{dev:results}
We show the dev.\ set results in~\cref{tab:dev}. Overall, the patterns of improvements hold across datasets and models. We base the test set result on the best-performing setups in the development set, i.e., \dwt{} and \adwt{}.

\section{Frequency Distribution of Skills}\label{app:skilldistribution}
We show the skill frequency distribution of the datasets in~\cref{fig:skilldistribution}, as mentioned in~\cref{subsec:longtail}. Here, we show evidence of the long-tail pattern in skills for each dataset. There is a cut-off at count 15 for \textsc{Green}, indicating that there are skills in the development set that occur more than 15 times.

\begin{table*}[t]
\begin{minipage}{\linewidth}
\centering
\resizebox{.75\linewidth}{!}{
    \begin{tabular}{|l|c|}
    \hline
    \multicolumn{2}{|c|}{JobBERTa $\rightarrow$ \textsc{SkillSpan}}\\\hline
    \hline
    Current token               & \texttt{IT}                    \\ \hline
    Gold label                  & \texttt{O} \\\hline
    LM prediction probs         & \texttt{[0.277, 0.404, 0.319]} \\
    \hline\hline
    Nearest neighbors ($k=8$)   & \texttt{['IT', 'Software', 'Software', 'Cloud', }\\ 
                                & \texttt{'Cloud', 'Database', 'Ag', 'software']} \\\hline
    Aggregated \knn{} scores    & \texttt{[0.000, 0.132, 0.868]}        \\\hline\hline

    Final predicted probs       & \texttt{[0.221, 0.350, 0.429]}\\\hline
    
    \end{tabular}}
    \caption{\textbf{Cherry Picked Qualitative Sample \nnose{} of Higher Precision.} We show a qualitative sample of using JobBERTa on \textsc{SkillSpan}. In this case, we see more weight being put on a specific tag, resulting in higher precision.}
    \label{tab:qualitative1}
\end{minipage}\hfill
\begin{minipage}{\linewidth}
    \centering
    \resizebox{.75\linewidth}{!}{
    \begin{tabular}{|l|c|}
    \hline
    \multicolumn{2}{|c|}{JobBERTa $\rightarrow$ \textsc{SkillSpan}}\\\hline
    \hline
    Current token               & \texttt{coding}                    \\ \hline
    Gold label                  & \texttt{B} \\\hline
    LM prediction probs         & \texttt{[0.988, 0.000, 0.012]} \\
    \hline\hline
    Nearest neighbors ($k=8$)   & \texttt{['programming', 'coding', 'programming', 'debugging', }\\ 
                                & \texttt{'scripting', 'writing', 'coding', 'programming']} \\\hline
    Aggregated \knn{} scores    & \texttt{[1.000, 0.000, 0.000]}        \\\hline\hline

    Final predicted probs       & \texttt{[0.991, 0.000, 0.009]}\\\hline\hline
    \cellcolor{gray!25}         &  \cellcolor{gray!25}        \\\hline\hline
    Current token               & \texttt{skills}                    \\ \hline
    Gold label                  & \texttt{I} \\\hline
    LM prediction probs         & \texttt{[0.000, 0.990, 0.010]} \\
    \hline\hline
    Nearest neighbors ($k=8$)   & \texttt{['skills', 'skills', 'skills', 'skills', 'skills', }\\ 
                                & \texttt{'skills', 'skills', 'skills']} \\\hline
    Aggregated \knn{} scores    & \texttt{[0.000, 1.000, 0.000]}        \\\hline\hline

    Final predicted probs       & \texttt{[0.000, 0.992, 0.008]}\\\hline
    
    \end{tabular}
    }
    \caption{\textbf{Cherry Picked Qualitative Sample \nnose{} of Multiple Tokens.} We show a qualitative sample of using JobBERTa on \textsc{SkillSpan} with multi-token annotations and how this behaves.}
    \label{tab:qualitative2}
\end{minipage}\hfill
\begin{minipage}{\linewidth}
    \centering
    \resizebox{.75\linewidth}{!}{
    \begin{tabular}{|l|c|}
    \hline
    \multicolumn{2}{|c|}{JobBERTa $\rightarrow$ \textsc{Green}}\\\hline
    \hline
    Current token               & \texttt{tools}                    \\ \hline
    Gold label                  & \texttt{I} \\\hline
    LM prediction probs         & \texttt{[0.250, 0.374, 0.379]} \\
    \hline\hline
    Nearest neighbors ($k=8$)   & \texttt{['tools', 'tools', 'transport', 'transport', }\\ 
                                & \texttt{'transport', 'transport', 'car', 'transport']} \\\hline
    Aggregated \knn{} scores    & \texttt{[0.124, 0.626, 0.250]}        \\\hline\hline

    Final predicted probs       & \texttt{[0.234, 0.399, 0.366]}\\\hline
    
    \end{tabular}
    }
    \caption{\textbf{Cherry Picked Qualitative Sample \nnose{} of Randomness.} We show a qualitative sample of using JobBERTa on \textsc{SkillSpan}.The language model puts high confidence on the tag \texttt{I}, which is the correct tag. Here the retrieved neighbors do not seem too relevant, which in this case is mostly random chance that it got it correctly.}
    \label{tab:qualitative3}
\end{minipage}\hfill
\begin{minipage}{\linewidth}
    \centering
    \resizebox{.75\linewidth}{!}{
    \begin{tabular}{|l|c|}
    \hline
    \multicolumn{2}{|c|}{JobBERTa $\rightarrow$ \textsc{SkillSpan}}\\\hline
    \hline
    Current token               & \texttt{optimistic}                    \\ \hline
    Gold label                  & \texttt{B} \\\hline
    LM prediction probs         & \texttt{[0.998, 0.000, 0.002]} \\
    \hline\hline
    Nearest neighbors ($k=8$)   & \texttt{['proactive', 'responsible', 'holistic', 'operational', }\\ 
                                & \texttt{'positive', 'open', 'professional', 'agile']} \\\hline
    Aggregated \knn{} scores    & \texttt{[1.000, 0.000, 0.000]}        \\\hline\hline

    Final predicted probs       & \texttt{[0.999, 0.000, 0.001]}\\\hline
    
    \end{tabular}
    }
    \caption{\textbf{Cherry Picked Qualitative Sample \nnose{} of Variety.} We show a qualitative sample of using JobBERTa on \textsc{SkillSpan}. The language model puts high confidence in the tag \texttt{B}, which is the correct tag. The retrieved neighbors are frequently relevant.}
    \label{tab:qualitative4}
\end{minipage}
\end{table*}

\section{Further Cross-dataset Analysis}\label{app:cr}
\paragraph{Precision and Recall Scores Cross-dataset.}
In~\cref{tab:pr}, we checked the precision and recall numbers for the cross-dataset setup with \adwt{} and JobBERTa as the backbone model. When using \nnose{}, we generally notice an increase in precision, with the largest when applied to \textsc{Sayfullina}. The largest gains are with respect to recall, we notice a significant gain in all setups, where the recall and precision increase is mixed. This indicates that \nnose{} is a useful method for both precision-focused and recall-focused applications, as we are storing skills in the datastore to be retrieved.

\section{Qualitative Results \nnose{}}\label{app:qualitative}

We show several qualitative results of \nnose{}. In \cref{tab:qualitative1}, we show a qualitative sample of using JobBERTa on \textsc{SkillSpan}. The current token is ``IT'' with gold label \texttt{O}. The language model puts 0.4 softmax probability on the tag \texttt{I}. By retrieving the nearest neighbors, the final probability mass gets shifted towards \texttt{O} with probability 0.43, which is the correct tag.

In \cref{tab:qualitative2}, we show a qualitative sample of using JobBERTa on \textsc{SkillSpan} with multi-token annotations and how this behaves. The current skill is ``coding skills'' with gold labels \texttt{B} and \texttt{I} respectively. Both the model and \knn{} puts high confidence in the correct label. Note that the nearest neighbors of ``coding'' are quite varied, which shows the benefit of \nnose{}. Note that all the retrieved ``skills'' tokens are from different contexts.

In \cref{tab:qualitative3}, we show a qualitative sample of using JobBERTa on \textsc{SkillSpan}. The current token is ``optimistic'' with gold label \texttt{B}. This is a so-called ``soft skill''. The language model puts high confidence in the tag \texttt{B}, which is the correct tag. The retrieved neighbors are frequently relevant, but sometimes less. This indicates that the retrieved neighbors (all soft skills) occur in similar contexts.

In \cref{tab:qualitative4}, we show a qualitative sample of using JobBERTa on \textsc{SkillSpan}. The current token is ``optimistic'' with gold label \texttt{B}. This is a so-called ``soft skill''. The language model puts high confidence in the tag \texttt{B}, which is the correct tag. The retrieved neighbors are frequently relevant, but sometimes less. This indicates that the retrieved neighbors (all soft skills) occur in similar contexts.

\end{document}